\def\year{2022}\relax
\documentclass[letterpaper]{article} 
\usepackage{aaai22}  
\usepackage{times}  
\usepackage{helvet}  
\usepackage{courier}  
\usepackage[hyphens]{url}  
\usepackage{graphicx} 
\usepackage{natbib}  
\usepackage{caption} 
\DeclareCaptionStyle{ruled}{labelfont=normalfont,labelsep=colon,strut=off} 
\usepackage{algorithm}
\usepackage{algorithmic}

\usepackage{newfloat}
\usepackage{listings}
\floatstyle{ruled}
\newfloat{listing}{tb}{lst}{}
\floatname{listing}{Listing}
\nocopyright
\usepackage{amsmath}
\usepackage{amssymb}
\DeclareMathOperator*{\argmax}{argmax}
\usepackage{multirow}
\usepackage{subfig}
\title{Variational Neural Temporal Point Process}
\author{
    Deokjun Eom\textsuperscript{\rm 1},
    Sehyun Lee\textsuperscript{\rm 1},
    Jaesik Choi\textsuperscript{\rm 1}\textsuperscript{\rm 2}
}
\affiliations{
    \textsuperscript{\rm 1}Graduate School of AI, Korea Advanced Institute of Science and Technology (KAIST) \thanks{This work was supported by Institute of Information \& communications Technology Planning \& Evaluation (IITP) grant funded by the Korea government(MSIT) (No.2019-0-00075, Artificial Intelligence Graduate School Program(KAIST))}\\
    \textsuperscript{\rm 2}INEEJI

    \{deokjun.eom, sehyun.lee, jaesik.choi\}@kaist.ac.kr
}

\begin{document}

\maketitle

\begin{abstract}
A temporal point process is a stochastic process that predicts which type of events is likely to happen and when the event will occur given a history of a sequence of events. There are various examples of occurrence dynamics in the daily life, and it is important to train the temporal dynamics and solve two different prediction problems, time and type predictions. Especially, deep neural network based models have outperformed the statistical models, such as Hawkes processes and Poisson processes. However, many existing approaches overfit to specific events, instead of learning and predicting various event types. Therefore, such approaches could not cope with the modified relationships between events and fail to predict the intensity functions of temporal point processes very well. In this paper, to solve these problems, we propose a variational neural temporal point process (VNTPP). We introduce the inference and the generative networks, and train a distribution of latent variable to deal with stochastic property on deep neural network. The intensity functions are computed using the distribution of latent variable so that we can predict event types and the arrival times of the events more accurately. We empirically demonstrate that our model can generalize the representations of various event types. Moreover, we show quantitatively and qualitatively that our model outperforms other deep neural network based models and statistical processes on synthetic and real-world datasets.

\end{abstract}

\section{Introduction}

Many different types of events occur irregularly in the world, and some events are strongly correlated with each other. There are also many examples of irregular sequences of the events. \cite{bacry2014hawkes} explained sequences of market prices based on the influence of market order arrivals using the multivariate Hawkes process. \cite{du2015dirichlet} used the Dirichlet Hawkes process on continuous-time document streams to predict document arrival time. \cite{choi2015constructing} proposed a multivariate context-sensitive Hawkes process for Electronic Health Records (EHR). In many cases, important information can be extracted from sequences of irregular past events, and to do so, modelling correlated structures among events is essential. The dependency in the occurrences of events is the key to inferring future events, and one of the most widely-used models for predicting future events is the temporal point process. 

The temporal point process is a stochastic process that characterizes the occurrences of asynchronous and discrete event sequences. The Poisson process and Hawkes process \cite{cox1980point, hawkes1971spectra, liniger2009multivariate, zhou2013learning} are statistical models used to learn the structures of irregular events. In Hawkes processes, specific kernel forms are used (e.g., an exponential kernel) and the goal is to train conditional intensity functions given past events. The Hawkes process maps sequences of past events to conditional intensity functions. The process is an example of a non-homogeneous Poisson process. Specifically, the intensity functions sum up all the effects of past events using fixed form of the kernels. In this way, the Hawkes processes can measure the influence of past events on a current event. However, Hawkes processes require a lot of computational time and are difficult to use for very long sequences of events.

Recently, many papers have employed deep neural networks to solve irregular event problems. Hawkes processes with recurrent neural network (RNN) \cite{du2016recurrent, mei2016neural, xiao2017modeling, li2018learning, huang2019recurrent} model the conditional intensity functions with RNN and can obtain deep representations of the history of event sequences. RNN models can learn the structures of irregular sequences of events, and the RNN based conditional intensity estimation outperforms statistical models, like Poisson processes or Hawkes processes. However, RNN have failed to detect long-term dependencies, and to overcome the limitations of RNN, \cite{vaswani2017attention} did not use a recurrent structure, and used self-attention layers to capture long-term dependencies. With this motivation, \cite{zhang2020self, zuo2020transformer} used self-attention or Transformer structures to train conditional intensity functions. Self-attention based Hawkes processes were effectively employed on a time-series dataset to capture dependency in history of event sequences, and the models achieved better performances than RNN based Hawkes processes.

However, existing approaches have only focused on integrating neural networks with Hawkes processes in deterministic way, and have failed to accurately predict intensity functions. The previous studies also suffer from overfitting problems, and some models are unable to predict various event types or easily converge to a few dominant events. This is because deterministic models are not appropriate for learning complex or noisy time series datasets which contain randomness and variability due to rare event types. In \cite{zuo2020transformer}, even though the model achieved state-of-the-art results on various datasets, they introduced two different linear layers to predict event type and time without using estimated intensity functions. In this paper, we propose a variational neural temporal point process (VNTPP) and use numerical approximations to obtain event time and type predictions using estimated intensity functions. First, we use inference and generative networks. The inference network predicts conditional intensity functions using the latent variable $z$. The generative network reproduces the input sequences of events and helps to learn the correlated structures of events and latent variables. Learning the distribution of the latent variable is helpful for predicting proper conditional intensity functions and various event types, thus preventing the overfitting problems that frequently happen in other models. 

In summary, this paper has four contributions:
\begin{itemize}
    \item This work suggests a novel multivariate temporal point process structure, the Variational Neural Temporal Point Process (VNTPP) with Transformer inference and generative networks. The intensity function is computed using the latent variable $z$, which improves the performances of the event and time predictions.
    \item We show quantitatively and qualitatively that the conditional intensity functions generated by VNTPP are more accurate than those from other models. In addition, we can estimate the mean and variance of the intensity functions by sampling the latent variables.
    \item We verify that the distribution of the latent variable $z$ is well trained using singular value decomposition. $z$'s are well distributed with respect to event types, without using any clustering methods.
    \item VNTPP is able to predict various event types and prevents the overfitting problems that exist in other neural network based temporal point processes.
\end{itemize}

\section{Background and Related Work}
\subsection{Multivariate Temporal Point Process}
A multivariate temporal point process is a stochastic process that characterizes the occurrences of asynchronous and discrete event sequences where the number of event types is more than 1. $\mathcal{S}=\{(k_i, t_i)\}_{i=1}^{L}$ is a event sequence where $(k_i, t_i)$ is the event type and timestamp of the event respectively. $\mathcal{H}_t:=\{(k^\prime, t^\prime)|t^\prime<t, k^\prime \in \mathcal{K}\}$ is a set of events that occur before time $t$ and $\mathcal{K}$ is a set of event types. In an infinitesimally short interval $[t, t+dt]$, we suppose that event type $k$ occurs at time $t$ with probability $\lambda_k(t)dt$ and $\lambda_k(t) \ge 0$ is considered to be the intensity function of event $k$ at time $t$. We can rewrite the intensity function as follows:
\begin{equation}
    \lambda_k(t)=\lim_{dt \to 0}\frac{P(\textrm{event }k\textrm{ occurs in }[t,t+dt)|\mathcal{H}_t)}{dt} 
\end{equation}
Therefore, the intensity function of all event types is as follows: 
\begin{equation}
    \lambda(t)=\sum_{k=1}^{|\mathcal{K}|}{\lambda_k(t)}.    
\end{equation}

The Hawkes process is an example of a multivariate temporal point process. The Hawkes process has a self-exciting structure, and the conditional intensity function is computed using the sum of the effects of history $\mathcal{H}_t$ and the base intensity of event $k$, $\mu_k$. The intensity function of event $k$ is given by
\begin{equation}
    \lambda_k(t) = \mu_k + \sum_{(k^\prime, t^\prime)\in \mathcal{H}_t}{\phi_{k,k^\prime}(t-t^\prime)},
\end{equation}
where the kernel function $\phi_{k,k^\prime}(t-t^\prime)>0$ represents the impact of event $k^\prime$ on event $k$. In \cite{pmlr-v31-zhou13a}, the kernel function can be represented by $\phi_{k,k^\prime}(t-t^\prime) = a_{k, k^\prime}f(t-t^\prime)$, where the coefficient $a_{k, k^\prime}$ means the mutual-exciting strength between event $k$ and $k^\prime$, and $f(t)$ is a time-decaying kernel. The base intensity $\mu_k>0$ is independent of the history.
The log-likelihood of the multivariate temporal point process over the observation interval $[0, T]$ is as follows:
\begin{equation}
    \mathcal{L}(\Lambda) = \sum_{i:t_i\le T}{\log \lambda_{k_i}(t_i)}-\int_{t=0}^{T}\lambda(t)dt,
\end{equation}
where $\mathcal{T}$ and $\mathcal{K}$ are sets of event timestamps and event types respectively, and $\Lambda=\{\lambda_{k_i}(t_i)| k_i \in \mathcal{K}, t_i \in \mathcal{T}\}$.

\begin{figure}[t!]
\centering
\includegraphics[width=0.40\textwidth]{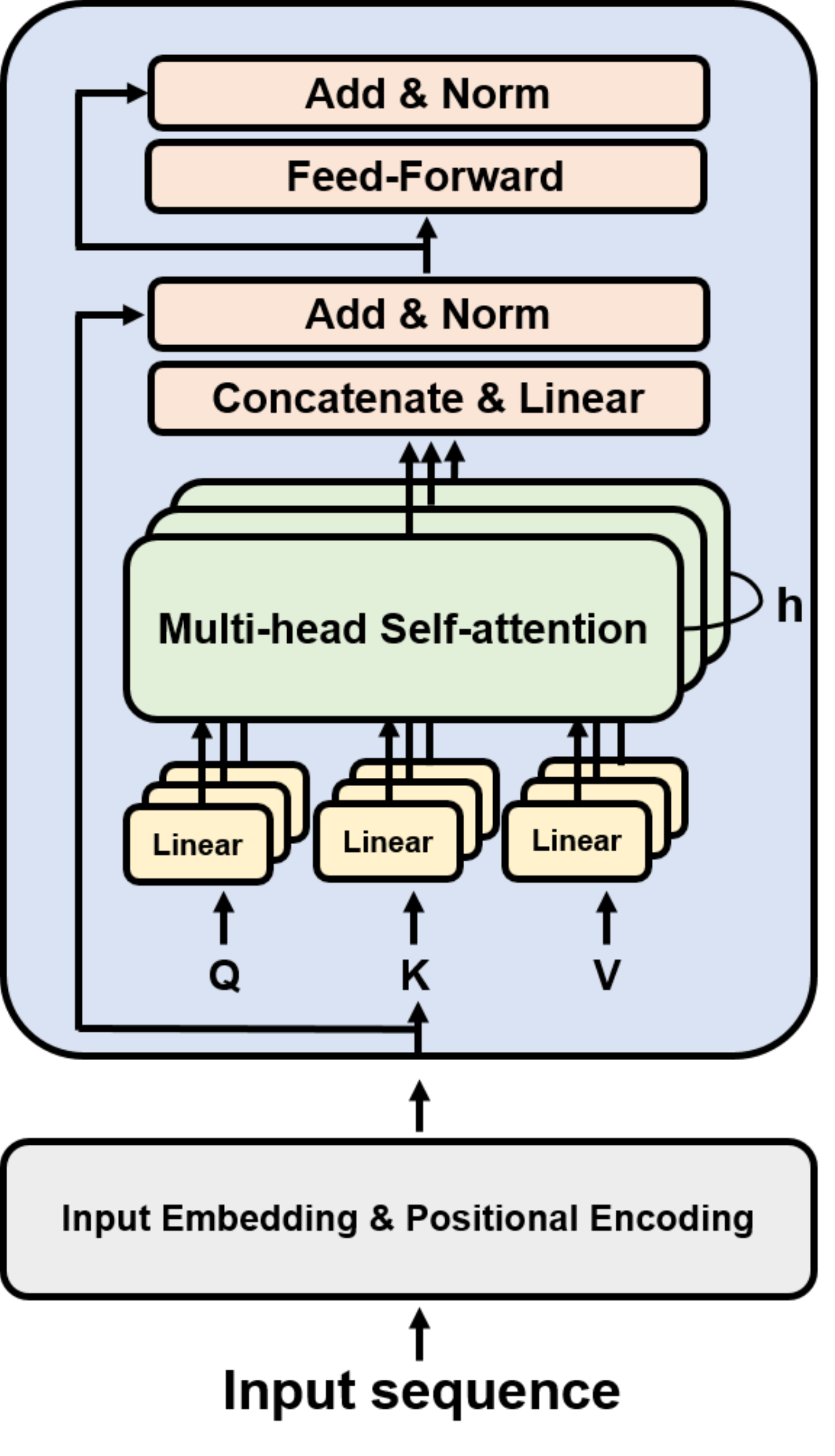}
\caption[Transformer encoder structure]{The detailed structure of Transformer encoder.}
\label{fig:transformer_architecture}
\end{figure}

\subsection{Variational Inference on Deep Neural Networks}
\cite{kingma2013auto} introduced variational inference learning algorithms for deep neural networks. In a case where the marginal likelihood $p_\theta(\textbf{x})$ is intractable, the true posterior distribution $p_\theta(\textbf{z}|\textbf{x})=p_\theta(\textbf{x}|\textbf{z})p_\theta(\textbf{z})/p_\theta(\textbf{x})$ is also intractable, where $\textbf{z}$ is an unobserved continuous latent variable and $\textbf{x}$ is a datapoint. To address this, they introduced an approximate posterior $q_\phi(\textbf{z}|\textbf{x})$ and optimized the variational lower bound using the approximate posterior, using a stochastic gradient descent estimator.

The marginal log-likelihood of all datapoints can be represented by the sum over the marginal log-likelihood of individual inputs as follows:
\begin{equation}
    \log p_\theta(\textbf{x}^{(1)},\dots \textbf{x}^{(N)})=\sum_{i=1}^{N}\log p_\theta(\textbf{x}^{(i)}).
\end{equation}
Instead of optimizing the marginal log-likelihood, they utilized the variational lower bound on marginal log-likelihood. The lower bound $\mathcal{L}(\theta,\phi;\textbf{x}^{(i)})$ can be written as follows:
\begin{equation}
\begin{split}
    \log p_\theta(\textbf{x}^{(i)}) &= \log \int p_\theta(\textbf{x}^{(i)}, \textbf{z}) d\textbf{z} \\
    &= \log \int p_\theta(\textbf{x}^{(i)}, \textbf{z}) \frac{q_\phi(\textbf{z}|\textbf{x}^{(i)})}{q_\phi(\textbf{z}|\textbf{x}^{(i)})} d\textbf{z}\\
    &\ge \mathbb{E}_{q_\phi(\textbf{z}|\textbf{x}^{(i)})}\bigg[\log \frac{ p_\theta(\textbf{x}^{(i)}, \textbf{z})}{q_\phi(\textbf{z}|\textbf{x}^{(i)})}\bigg]\\
    &=\mathbb{E}_{q_\phi(\textbf{z}|\textbf{x}^{(i)})}\bigg[-\log q_\phi(\textbf{z}|\textbf{x}^{(i)})+\log p_\theta(\textbf{x}^{(i)},\textbf{z})\bigg] \\
    &=\mathcal{L}(\theta,\phi;\textbf{x}^{(i)}), \label{eq:elbo}
\end{split}
\end{equation}
where the inequality comes from Jensen's inequality.

The proposed approach in this paper uses the latent variable $\textbf{z}$ to make conditional intensity functions. Using the predicted intensity functions, we can determine estimates of event type and time. The motivation for our approach is from classification models using variational approximations. \cite{kingma2014semi, maaloe2016auxiliary, willetts2020semi, xie2021semisupervised} utilized various variational inference structures on semi-supervised classification tasks. They showed that variational approaches improve the performance for classification tasks, and one of the tasks in temporal point processes is event classification. In addition, \cite{chung2015recurrent} showed that deterministic structures are not suitable for modelling the variability or randomness of complex time-series datasets. Therefore, we suggest a novel approach of neural temporal point processes utilizing variational inference.

\subsection{Transformer}
Our proposed model contains a Transformer \cite{vaswani2017attention} structure on inference and generative networks. Transformer has been used in various types of sequential datasets, including natural language processing \cite{devlin2018bert, radford2019language} and time-series predictions \cite{wu2020deep}. The Transformer consists of multi-head self-attentions with position-wise fully connected networks. The multi-head attention learns different representations from different attention heads. The simplified structure of multi-head attention is as follows:
\begin{align}
    &\textrm{Attention}(Q,K,V)=\textrm{softmax}(\frac{QK^T}{\sqrt{d_k}})V, \nonumber\\
    &\textrm{head}_i = \textrm{Attention}(QW_i^Q,KW_i^K,VW_i^V) \label{eq:1},\\
    &\textrm{MultiHead}(Q,K,V) =   \textrm{Concat}(\textrm{head}_1,\dots,\textrm{head}_h)W^O,  \nonumber
\end{align}
where $K,Q,V$ are key, query, and value matrices respectively, and $\sqrt{d_k}$ is the dimension of the keys and queries. The one of the advantages in this self-attention structure is that each hidden representation can attend all the hidden vectors of previous layers. Therefore, we can solve long-term dependency problem that exists in RNN based models. Fig \ref{fig:transformer_architecture} describes the detailed structure of Transformer encoder.

In multivariate temporal point processes, \cite{zhang2020self, zuo2020transformer} applied self-attentive and Transformer structures to train intensity functions. However, the models were easily overfitted and in \cite{zuo2020transformer}, they trained two additional linear layers to predict event type and time. In this paper, we show qualitatively and quantitatively that the variational structure helps to estimate the intensity functions.

\section{Variational Neural Temporal Point Process}

\begin{figure*}[t]%
    \centering
    \subfloat[\centering ]{{\includegraphics[width=0.7\textwidth]{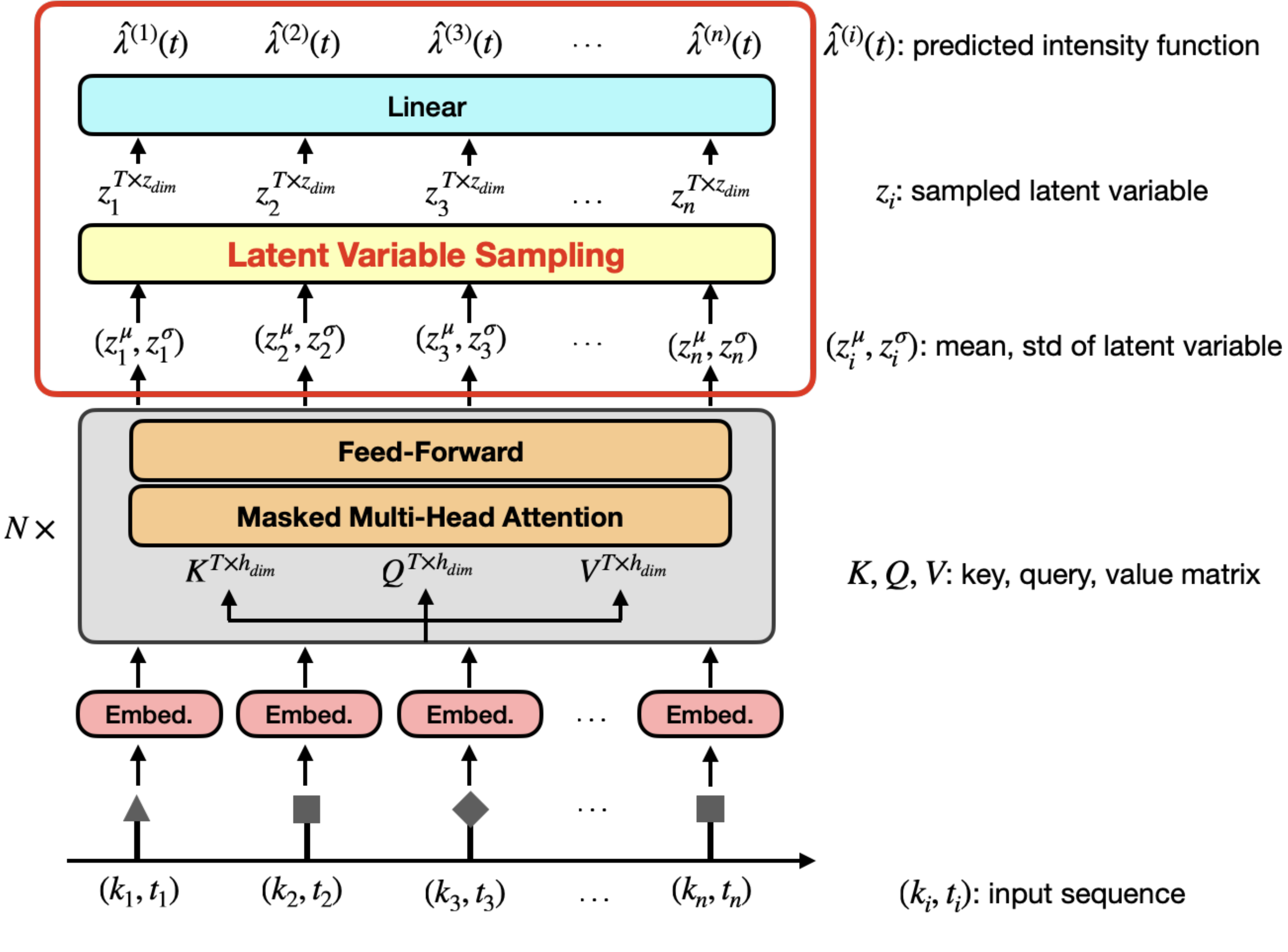} }}%
    \subfloat[\centering ]{{\includegraphics[width=0.28\textwidth]{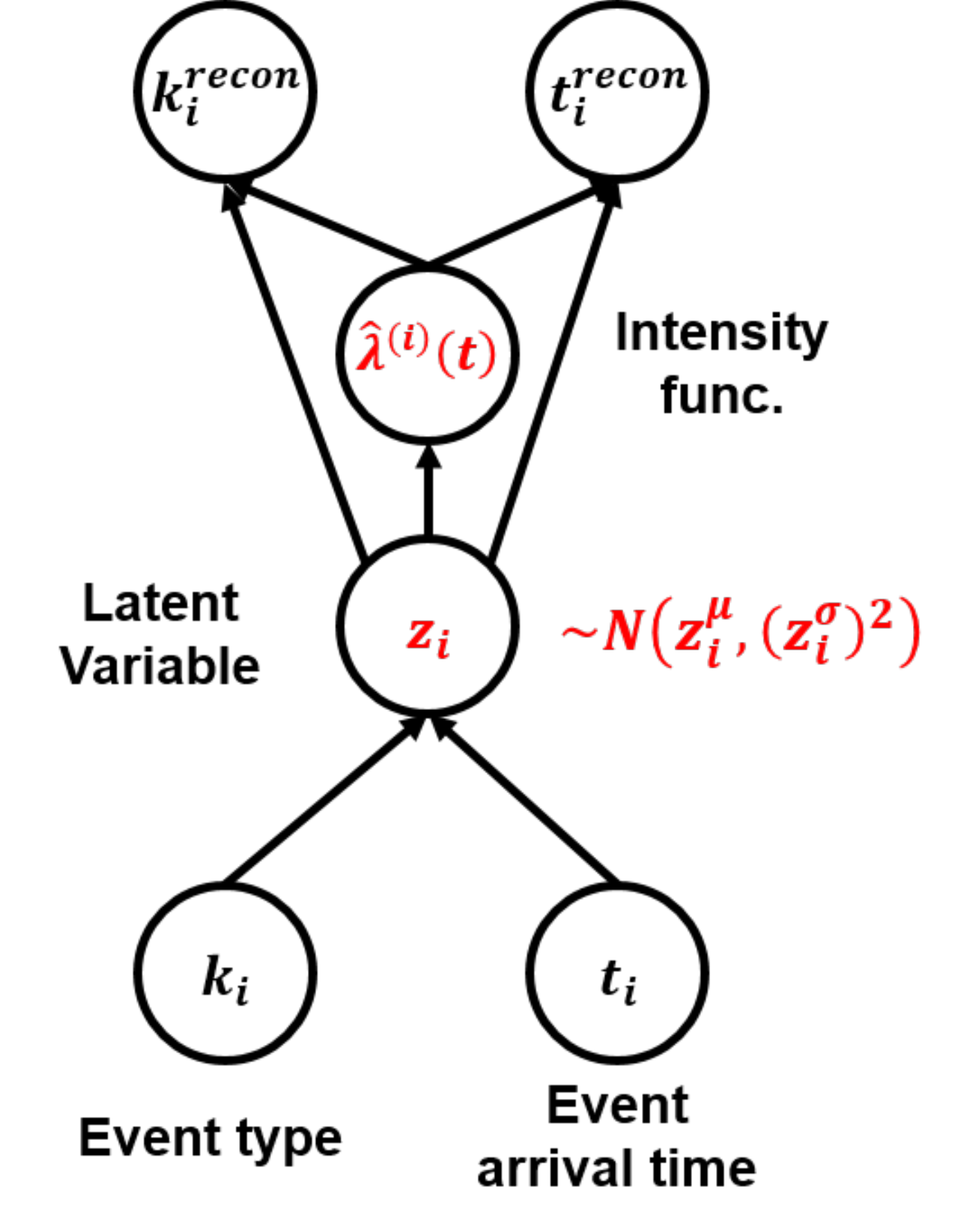} }}%
    \caption[VNTPP structure]{The left figure describes the VNTPP architecture used to compute the latent variable $z$ and the conditional intensity function $\lambda$. The gray box represents the inference network used to compute the mean and variance of latent variable $z$. The latent variable $z$ is used to determine the conditional intensity function $\lambda$. Additionally, by sampling the latent variable repeatedly, we can estimate the distributions of the intensity functions. $z$ and $\lambda$ are utilized to reconstruct the input variables. The right figure shows a graphical structure of VNTPP.}
    \label{fig:model_architecture}%
\end{figure*}

\subsection{Inference and Generative Networks}
We use Transformer structures for the inference and generative networks. In this section, we explain the detailed structures of the Transformer. Since a Transformer does not contain recurrent structures, positional information needs to be given in the model. For this purpose, \cite{vaswani2017attention} introduced a positional embedding function. It consists of sine and cosine function depending on the positions of the input sequences. \cite{zuo2020transformer} applied the temporal encoding on timestamps as follows:
\begin{align}
    \textrm{PE}(t, l) = 
    \begin{cases}
    \cos(t/10000^{(l-1)/D}), & \text{if } l \text{ is odd}\\
    \sin(t/10000^{l/D}), & \text{if } l \text{ is even}\\
    \end{cases}
\end{align}
where $D$ is the number of hidden dimension in the model. Likewise, we use an embedding layer to encode event types. The encoded representations of event type and time can be represented as follows:
\begin{align}
    \mathbf{h}_i = \text{U}\mathbf{k}_i + [\text{PE}(t_i, 0), \dots, \text{PE}(t_i,D-1)]^\top,
\end{align}
where $\text{U}\in \mathbb{R}^{D, K}$ is an embedding matrix for event type, and $\mathbf{k}_i$ is one-hot encoding of an event. We use $\mathbf{h}_i$ as an input for the Transformer decoder, as described in \cite{vaswani2017attention}.

In the inference network, the latent variable $\textbf{z}$ is sampled from approximate posterior distribution. Two separate linear layers produce the mean and variance of $\textbf{z}$ using the outputs of the Transformer in the inference network. To train the mean and variance of $\textbf{z}$ with stochastic gradient descent, we use reparameterization trick described in \cite{kingma2013auto}. Reparameterization trick have been widely used in variational inference on deep neural networks \cite{kingma2013auto, kingma2015variational}. If the latent variable $\textbf{z}$ is sampled directly from the Gaussian distribution with a mean $\boldsymbol{\mu}_z$ and $\boldsymbol{\sigma}_z$ without the reparameterization trick, it is impossible to back-propagate the gradients to the inference network, and the parameters of the inference network cannot be updated. The reparameterization trick is as follows:
\begin{align}
    \mathbf{z} = \boldsymbol{\mu}_z + \boldsymbol{\sigma}_z \circ \boldsymbol{\epsilon},
\end{align}
where $\boldsymbol{\mu}_z$ and $\boldsymbol{\sigma}_z$ are the mean and standard deviation of latent variable $\mathbf{z}$, and $\boldsymbol{\epsilon}$ is sampled from the i.i.d. standard normal distribution $\mathbf{z} \sim \mathcal{N}(\boldsymbol{0},\textbf{I})$. The symbol $\circ$ is element-wise multiplication.

To obtain a positive intensity function, we use a linear layer with a softplus activation function. 
\begin{align}
\begin{split}
    \lambda_{k_i}(t_i) &= \text{softplus}(\beta_{k_i} + \textbf{w}_{k_i}^\top \textbf{z} - \alpha_{k_i} \cdot (t_i-t_{i-1})), \\
    \lambda_{k_i}(t_i) &= \text{softplus}(\beta_{k_i} + \textbf{w}_{k_i}^\top \textbf{z} + \exp (- \alpha_{k_i} \cdot (t_i-t_{i-1}))),    
\end{split}
\end{align}
where $\textbf{z}$ is latent variable sampled from an approximate posterior. The first case is a linearly decaying (VNTPP-L), and the second case is an exponentially decaying (VNTPP-E) with respect to time. The softplus function is more stable than ReLU since softplus is differentiable on a real line, and the derivative is always non-zero (greater than zero). The latent variable $\textbf{z}$ is effective when predicting intensity functions. $\alpha_k$ controls the decreasing or increasing properties with respect to time. In addition, $\alpha_k$ is also a trainable parameter, and different for each event. $\beta_k$ has a role in measuring the base intensity depending on event type. We can obtain the conditional intensity functions for each event type and the intensity functions are used to predict the next event type and time using numerical approximation.

\subsection{Log-likelihood}
We combine the variational lower bound and point process log-likelihood. We replace $\textbf{x}^{(i)}$ in $\mathcal{L}(\theta,\phi;\textbf{x}^{(i)})$ from \ref{eq:elbo} with history $\mathcal{H}$ and derive the variational lower bound. The derivation of the lower bound is shown below:
\begin{align}
\begin{split}
    \log p(\mathcal{H}) &= \log \int q(\textbf{z}|\mathcal{H})\frac{p(\mathcal{H}, \textbf{z})}{q(\textbf{z}|\mathcal{H})}d\textbf{z} \\
    &\ge \int q(\textbf{z}|\mathcal{H}) \log \frac{p(\mathcal{H}, \textbf{z})}{q(\textbf{z}|\mathcal{H})} d\textbf{z} \\
    &= \int q(\textbf{z}|\mathcal{H}) \log \frac{p(\textbf{z})p(\mathcal{H}|\textbf{z})}{q(\textbf{z}|\mathcal{H})} d\textbf{z} \\
    &= \mathbb{E}_{q(\textbf{z}|\mathcal{H})}\big[\log p(\mathcal{H}|\textbf{z})\big] - \textrm{KL}\big[q(\textbf{z}|\mathcal{H})||p(\textbf{z})\big],
\end{split}
\end{align}
where the inequality can be obtained from Jensen's inequality. We suppose $p(\textbf{z})$ follows the unit Gaussian distribution, and $q(\textbf{z}|\mathcal{H})$ also follows the Gaussian distribution with mean $\boldsymbol{\mu}_\textbf{z}$ and variance $\boldsymbol{\sigma}_\textbf{z}^2$. Therefore, we can compute the KL divergence analytically as follows:
\begin{align}
    \textrm{KL}\big[q(\textbf{z}|\mathcal{H})||p(\textbf{z})\big] = -\frac{1}{2}\sum_{j=1}^{J}\bigg(1+\log \sigma_{\textbf{z},j}^2-\mu_{\textbf{z},j}^2-\sigma_{\textbf{z},j}^2 \bigg),
\end{align}

\begin{figure*}[t!]
\centering
\includegraphics[width=0.99\textwidth]{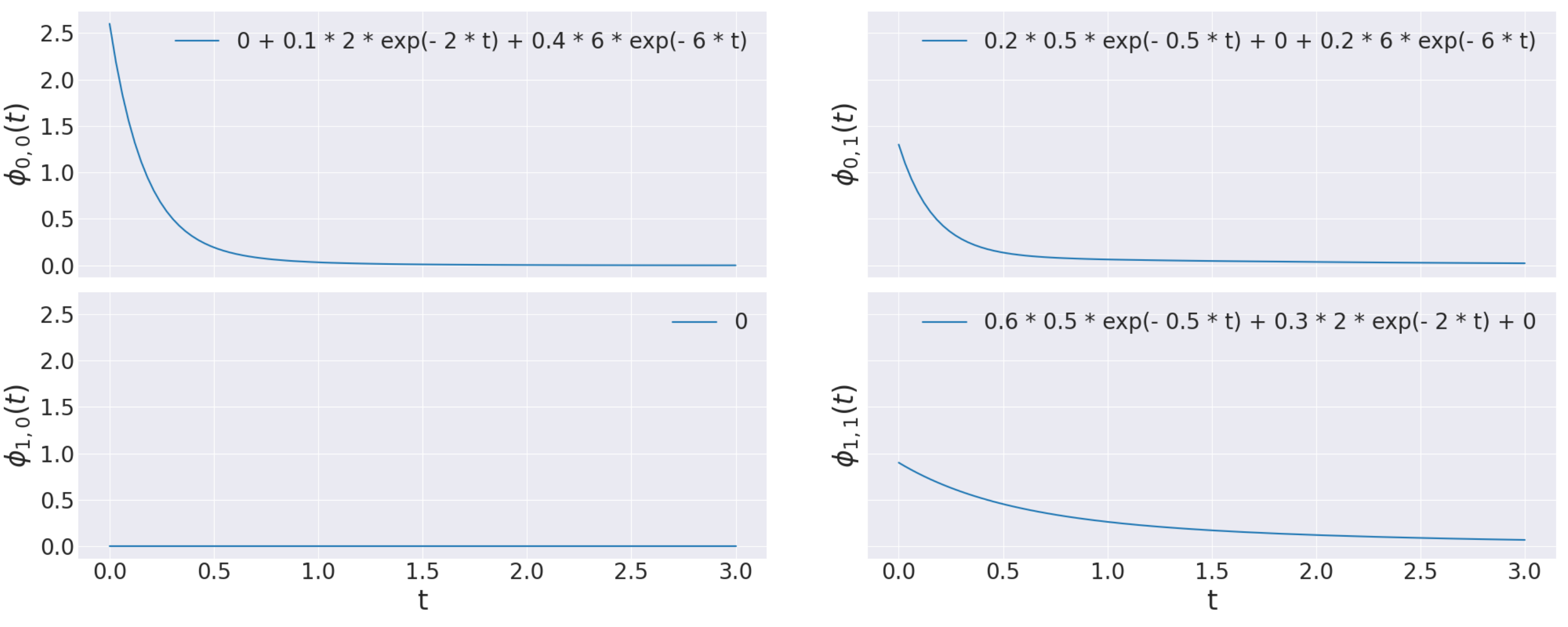}
\caption[The kernels of synthetic 1 dataset]{The kernels of synthetic 1 dataset. The kernel function $\phi_{i,j}(t)$ represents the impact of event $j$ on event $i$. }
\label{fig:synthetic_1_kernel}
\end{figure*}

\begin{figure*}[t!]
\centering
\includegraphics[width=0.99\textwidth]{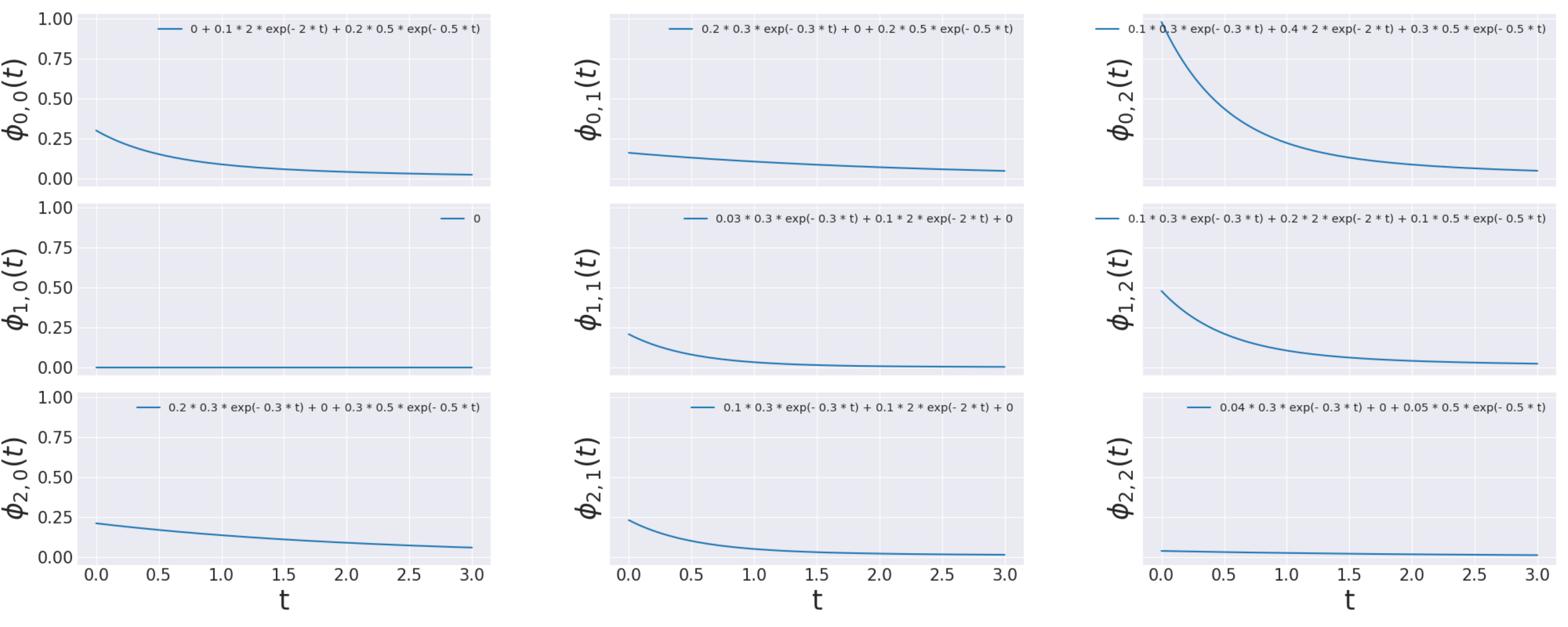}
\caption[The kernels of synthetic 2 dataset]{The kernels of synthetic 2 dataset. The kernel function $\phi_{i,j}(t)$ represents the impact of event $j$ on event $i$.}
\label{fig:synthetic_2_kernel}
\end{figure*}

\begin{table*}[h]
\centering
\begin{tabular}{cccccc} 
\hline
      & MIMIC-II & StackOverflow & Retweet & Synthetic 1 & Synthetic 2  \\ 
\hline
HP-EK        &  \textbf{0.841 $\pm$ 0.003}   & 0.437 $\pm$ 0.001  & 0.464 $\pm$ 0.000 & \textbf{0.632 $\pm$ 0.000} & \textbf{0.495 $\pm$ 0.000} \\
HP-GK        & \textbf{0.843 $\pm$ 0.000} & 0.441 $\pm$ 0.003 & -  &  0.619 $\pm$ 0.000  & \textbf{0.495 $\pm$ 0.000} \\
SAHP         & 0.401 $\pm$ 0.000 & 0.330 $\pm$ 0.013 & 0.512 $\pm$ 0.036 & 0.620 $\pm$ 0.000 & 0.361 $\pm$ 0.000 \\
THP          & 0.401 $\pm$ 0.000 & 0.458 $\pm$ 0.000 & 0.503 $\pm$ 0.009 & 0.629 $\pm$ 0.000 & 0.494 $\pm$ 0.000 \\
VNTPP-L (ours) & \textbf{0.843 $\pm$ 0.000} & \textbf{0.462 $\pm$ 0.000} & \textbf{0.596 $\pm$ 0.006} & \textbf{0.631 $\pm$ 0.002}  & \textbf{0.495 $\pm$ 0.000}\\
VNTPP-E (ours) & \textbf{0.843 $\pm$ 0.008}  & 0.461 $\pm$ 0.000 & 0.586 $\pm$ 0.008 & \textbf{0.632 $\pm$ 0.001}  & \textbf{0.495 $\pm$ 0.000}\\
\hline
\end{tabular}
\caption[F1 score of event type prediction]{F1 score of event type prediction}
\label{tab:f1_result}
\end{table*}

\begin{table*}[h]
\centering
\begin{tabular}{cccccc} 
\hline
& MIMIC-II & StackOverflow & Retweet ($\times 10^4$) & Synthetic 1 & Synthetic 2  \\ 
\hline
HP-EK        & 0.938 $\pm$ 0.075 & 1.063 $\pm$ 0.009 & 1.665 $\pm$ 0.000 & 1.992 $\pm$ 0.006 & 1.651 $\pm$ 0.006 \\
HP-GK        & 0.912 $\pm$ 0.049 & 1.079 $\pm$ 0.002 &   -   & 1.960 $\pm$ 0.002 & 1.659 $\pm$ 0.016 \\
SAHP         & \textbf{0.837 $\pm$ 0.010} & 43.847 $\pm$ 2.762 & 1.663 $\pm$ 0.002 & 2.178 $\pm$ 0.069 & 1.786 $\pm$ 0.032 \\
THP          & 0.890 $\pm$ 0.004 & 0.986 $\pm$ 0.005 & 1.627 $\pm$ 0.001 & 1.842 $\pm$ 0.002 & 1.532 $\pm$ 0.001 \\
VNTPP-L (ours) & \textbf{0.838 $\pm$ 0.007} & \textbf{0.964 $\pm$ 0.002}  & 1.634 $\pm$ 0.012 & 1.848 $\pm$ 0.001 & \textbf{1.523 $\pm$ 0.003} \\
VNTPP-E (ours) & 0.881 $\pm$ 0.015  & 0.973 $\pm$ 0.004 & \textbf{1.456 $\pm$ 0.016} & \textbf{1.832 $\pm$ 0.001} & \textbf{1.520 $\pm$ 0.003}  \\
\hline
\end{tabular}
\caption[RMSE of event time prediction]{RMSE of event time prediction}
\label{tab:rmse_result}
\end{table*}

\begin{table*}[t!]
\centering
\begin{tabular}{ccccc} 
\hline
             & \multicolumn{2}{c}{Synthetic 1} & \multicolumn{2}{c}{Synthetic 2}  \\
             & RMSE   & MAE                    & RMSE   & MAE                     \\ 
\hline
SAHP         & 5.315 $\pm$ 0.167 &  4.045 $\pm$  0.096 & 2.423 $\pm$ 0.117& 1.684 $\pm$  0.045    \\
THP          & 4.917 $\pm$ 0.030 &  2.698 $\pm$ 0.017  & 3.782 $\pm$ 0.017 & 1.979 $\pm$ 0.005     \\
VNTPP-L  & \textbf{2.709 $\pm$ 0.022} & \textbf{2.113 $\pm$ 0.016} & 0.587 $\pm$ 0.038 & 0.454 $\pm$ 0.015\\
VNTPP-E  & 2.837 $\pm$ 0.010 & 2.165 $\pm$ 0.010 & \textbf{0.395 $\pm$ 0.070} & \textbf{0.245 $\pm$ 0.015}\\
\hline
\end{tabular}
\caption[Intensity function estimation results]{Intensity function estimation results. VNTPP outperforms state-of-the-art baselines.}
\label{tab:intensity_result}
\end{table*}

where $J$ is the number of dimensions of the latent variable $\textbf{z}$. The term $ \mathbb{E}_{q(\textbf{z}|\mathcal{H})}\big[\log p(\mathcal{H}|\textbf{z})\big]$ evaluates reconstructions of the generative network. The generative network reconstructs event type and time at the same time with two separate linear layers. The input of the two linear layers is the Transformer output of the generative network. Latent variable $\textbf{z}_{t_i}$ is a latent variable of the history $\mathcal{H}_{t_i}$. The final objective function to be optimized is as follows:
\begin{align}
    \mathcal{L}(\theta, \phi;  \mathcal{H}, \Lambda) = \sum_{i:t_i\leq T}\bigg[\mathbb{E}_{q(\textbf{z}_{t_i}|\mathcal{H}_{t_i})}\big[\log p(\mathcal{H}_{t_i}|\textbf{z}_{t_i})\big] \nonumber\\
    - \textrm{KL}\big[q(\textbf{z}_{t_i}|\mathcal{H}_{t_i})||p(\textbf{z}_{t_i})\big]
    + \log\lambda_{k_i}(t_i)\bigg]-\int_{t=0}^{T}\lambda(t)dt. \label{eq:log-likelihood}
\end{align}
where $\mathcal{T}$ and $\mathcal{K}$ are sets of event timestamps and event types respectively, $\mathcal{H}=\{\mathcal{H}_{t_i}|t_i \in \mathcal{T} \}$, and $\Lambda=\{\lambda_{k_i}(t_i)| k_i \in \mathcal{K}, t_i \in \mathcal{T}\}$.

\begin{figure*}[t!]%
    \centering
    \subfloat[\centering  ]{{\includegraphics[width=0.48\textwidth]{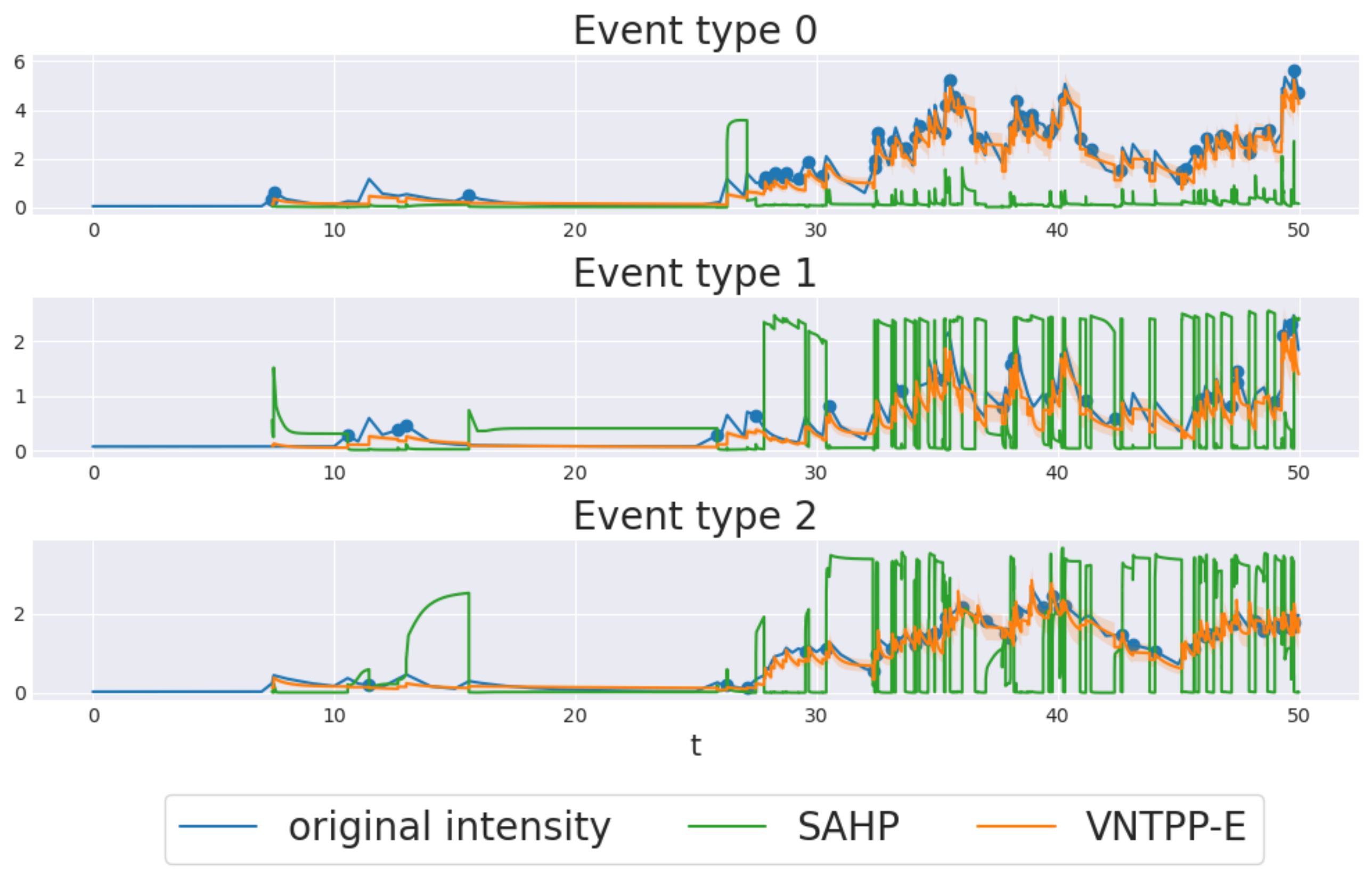} }}%
    \:
    \subfloat[\centering ]{{\includegraphics[width=0.48\textwidth]{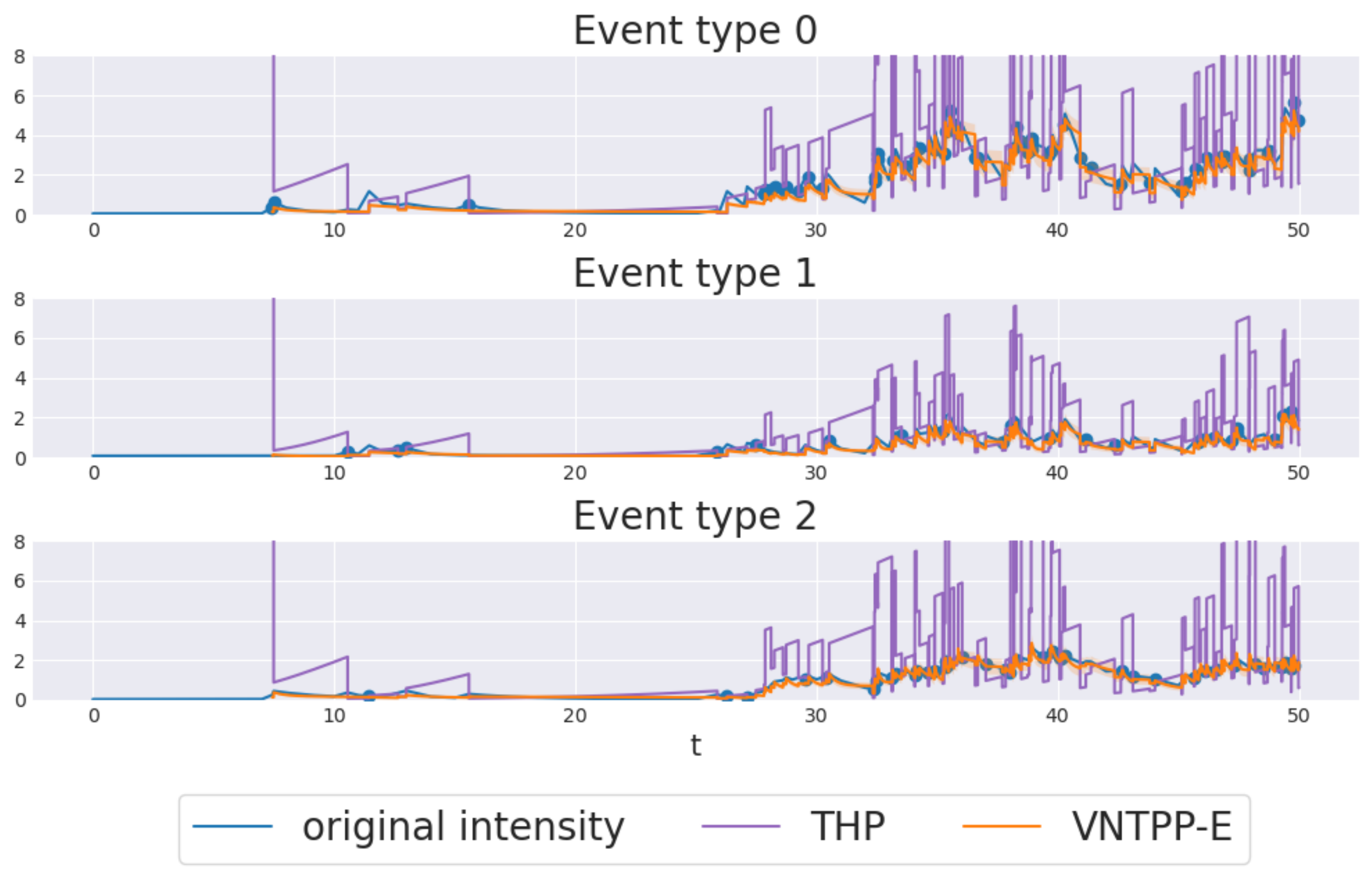} }}%
    \caption[Intensity function prediction]{Intensity function estimation example on Synthetic 2 dataset. The Synthetic 2 dataset consists of three types of event, and accordingly we have three original intensity functions for each event type. A blue dot indicates the occurrence of the event. The estimated intensity functions of VNTPP are found to be closer to the original intensity functions than SAHP and THP.}%
    \label{fig:intensity_function}
\end{figure*}

\begin{figure}[t!]%
    \centering
    \subfloat[\centering ]{{\includegraphics[width=0.22\textwidth]{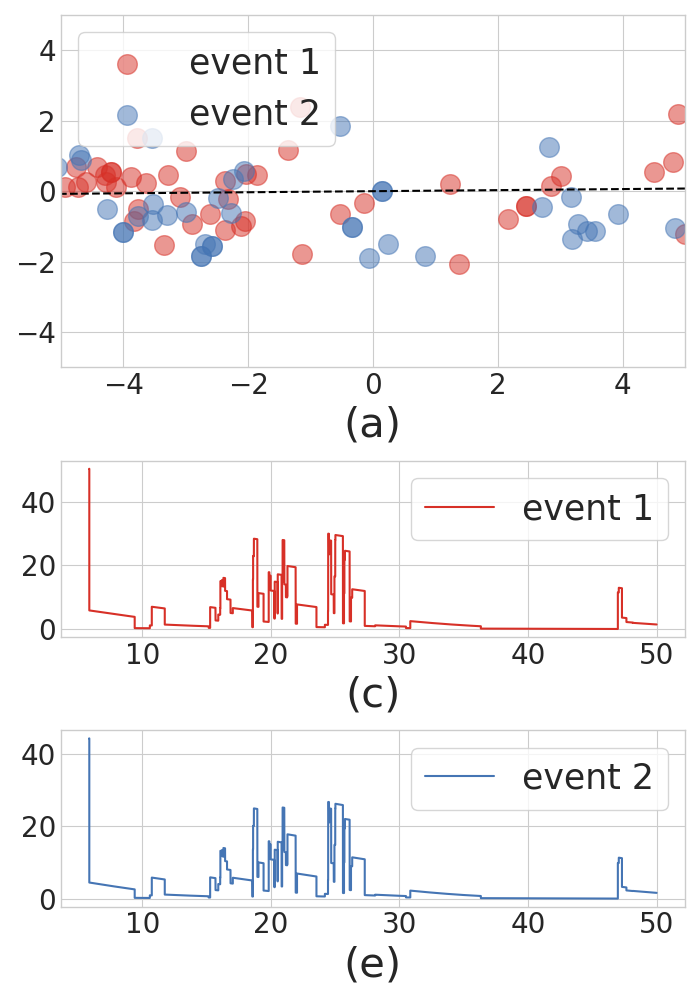} }}%
    \:
    \subfloat[\centering ]{{\includegraphics[width=0.22\textwidth]{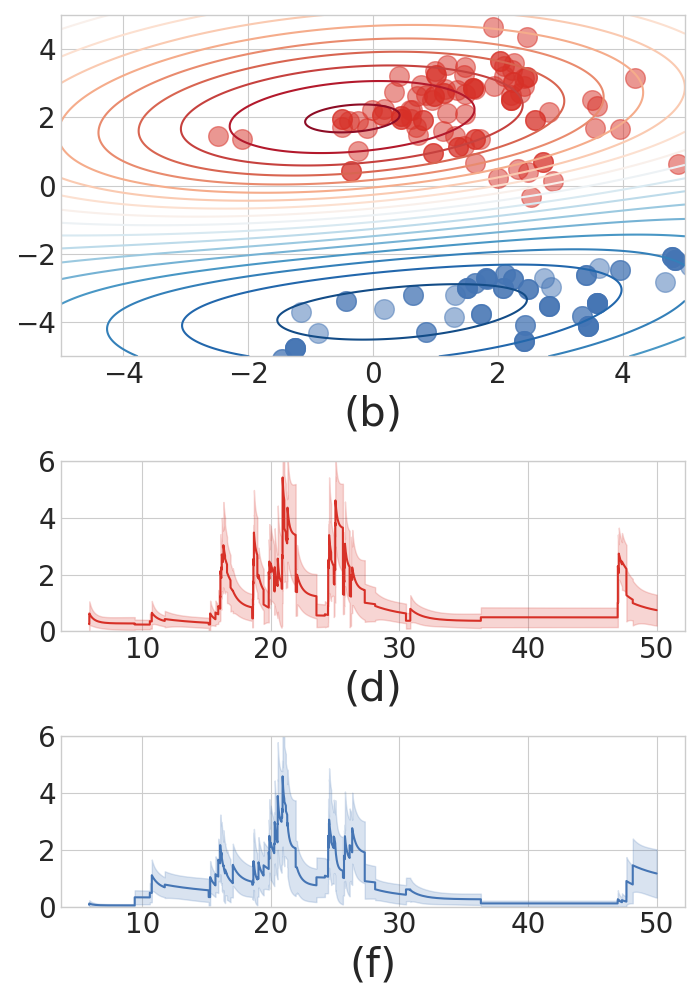} }}%
    \caption[Difference between deterministic model and stochastic model]{These figures describe the difference between deterministic model (left, THP) and stochastic model (right, VNTPP). The left figures show point estimation results. On the other hand, in right figures, we can estimate the distribution of the intensity function so that VNTPP can deal with the data uncertainty by sampling the latent variables. Figure (a) and (b) describe the latent space and imply that VNTPP generalizes the representation of latent space better than THP. The intensity functions are shown in figure (c)-(f).}%
    \label{fig:density_contour}%
\end{figure}

\subsection{Event and Time Prediction}
The probability density function of event time given the history $\mathcal{H}_{t_{i+1}}=\{(k_1,t_1),(k_2, t_2) \dots, (k_i, t_i) \}$ is as follows:
\begin{align}
    f_{i+1}(t) = p(t_{i+1}=t>&t_i|\mathcal{H}_{t_{i+1}}) 
    =  \nonumber\\
    &\lambda(t)\exp\bigg(-\int_{t_i}^{t}\lambda(u)du \bigg), \label{eq:pdf_event_time}
\end{align}
where $\lambda(t)=\sum_{k=1}^{|\mathcal{K}|}{\lambda_k(t)}$. To predict the event time, we compute the expectation of event time as follows:
\begin{equation}
    \hat{t}_{i+1} = \int_{t_i}^{\infty} t f_{i+1}(t)dt.
    \label{eq:time_pred}
\end{equation}
In addition, the probability of event type $k_{i+1}$ given event time $t_{i+1}$ is equal to:
\begin{equation}
    p(k_{i+1} = k | t_{i+1}) = \frac{\lambda_{k}(t_{i+1})}{\lambda(t_{i+1})}.
\end{equation}
Instead of using conditional probability, we use the marginal probability of the event type since we do not know the next event time when predicting the next event type:
\begin{equation}
    p(k_{i+1} = k) = \int_{t_i}^{\infty}\frac{\lambda_{k}(t)}{\lambda(t)} f_{i+1}(t)dt.
    \label{eq:marginal_event_type}
\end{equation}
Using this marginal probability of the event type, we select an event that has a maximum probability as the predicted next event type.
\begin{equation}
    \hat{k}_{i+1} =\argmax_{k \in \mathcal{K}} p(k_{i+1} = k).
\end{equation}
The entire structure of the VNTPP is shown in Fig \ref{fig:model_architecture}.

\subsection{Numerical Approximation}
In this framework, we should compute several integrals to obtain the log-likelihood in equation \ref{eq:log-likelihood}, the probability density function (pdf) of event time in equation \ref{eq:pdf_event_time}, the expectation of event time pdf in equation \ref{eq:time_pred}, and the marginal probability of the event type in equation \ref{eq:marginal_event_type}. First, we use Monte Carlo integration to compute $\int_{t=0}^{T}\lambda(t)dt$ as follows:
\begin{align}
    \int_{t=0}^{T}\lambda(t)dt \approx  \sum_{i=1}^{L-1}(t_{i+1}-t_i)\bigg(\frac{1}{M}\sum_{m=1}^{M}\lambda\Big((t_{i+1}-t_i)u_m+t_i\Big)    \bigg), 
\end{align}
where  $u_m$'s are sampled from $\text{Unif}(0,1)$ and $M$ is the number of samples for each timestamp. In the other integrals, we use right Riemann sum as follows:
\begin{align}
    \int_0^{T}f(t)dt  \approx \sum_{i=1}^n f(t_i) \Delta t_i,
\end{align}
where $0=t_0<t_1<\cdots<t_n=T$ and $\Delta t_i = t_i - t_{i-1}$. Another method is the Trapezoidal sum, and the integrals can be obtained by averaging the left and right Riemann sums:
\begin{align}
    \int_0^{T}f(t)dt  \approx \sum_{i=1}^n \frac{f(t_i)+f(t_{i-1})}{2} \Delta t_i.
\end{align}
There is very little difference between the right Riemann sum and the Trapezoidal sum in our experiments.

\section{Experimental Evaluations}

\subsection{Datasets}
We use three real-world datasets and two synthetic datasets. The synthetic datasets are generated under fixed kernels using the open-source Python library \textit{tick} \cite{bacry2017tick}. The detailed description of the datasets and hyperparameter settings of VNTPP can be found in Appendix

\begin{itemize}
    \item \textbf{MIMIC-II} \cite{lee2011open} MIMIC-II (Multiparameter Intelligent Monitoring in Intensive Care) dataset is a collection of clinical visits of patients for seven years. One record consists of a sequence of diagnosis and timestamps for each visit. The total types of diagnosis are 75 and the dataset has 650 sequences in total. The mean of the sequence lengths is 3.72.
    \item \textbf{StackOverflow} \cite{leskovec2014snap} StackOverflow has a reward system and there are 22 types of rewards in this dataset. Users can be awarded by answering posted questions on the StackOverflow website. An input is a sequence of award records of a user. The dataset has 6,633 sequences in total. The mean of the sequence lengths is 72.43.
    \item \textbf{Retweet} \cite{zhao2015seismic} There are three categories of user types, and the user types depend on the number of followers. An input is a sequence of tweets of various users, and there is a user type for each tweet. The dataset has 24,000 sequences in total and the number of event types is 3. The mean of the sequence lengths is 108.75.
    \item \textbf{Synthetic 1} $\:$ The number of events in this dataset is 2, and the dataset is generated using the Hawkes process with exponential kernels. The dataset has 9,960 sequences in total. The mean of the sequence lengths is 58.47. The kernels are described in Fig \ref{fig:synthetic_1_kernel}
    \item \textbf{Synthetic 2} $\:$ The number of events in this dataset is 3, and the dataset is generated using the Hawkes process with exponential kernels. The dataset has 9,868 sequences in total. The mean of the sequence lengths is 65.71. The kernels are described in Fig \ref{fig:synthetic_2_kernel}
\end{itemize}

\subsection{Baselines}
We select two Hawkes processes with different kernels and two state-of-the art neural network based models.
\begin{itemize}
    \item Hawkes process with exponential kernels (HP-EK)
    \item Hawkes process with Gaussian kernels (HP-GK)
    \item Self-attentive Hawkes process \cite{zhang2020self} (SAHP)
    \item Transformer Hawkes process \cite{zuo2020transformer} (THP)
\end{itemize}

\begin{figure}[t!]%
    \centering
    \subfloat[\centering ]{{\includegraphics[width=0.22\textwidth]{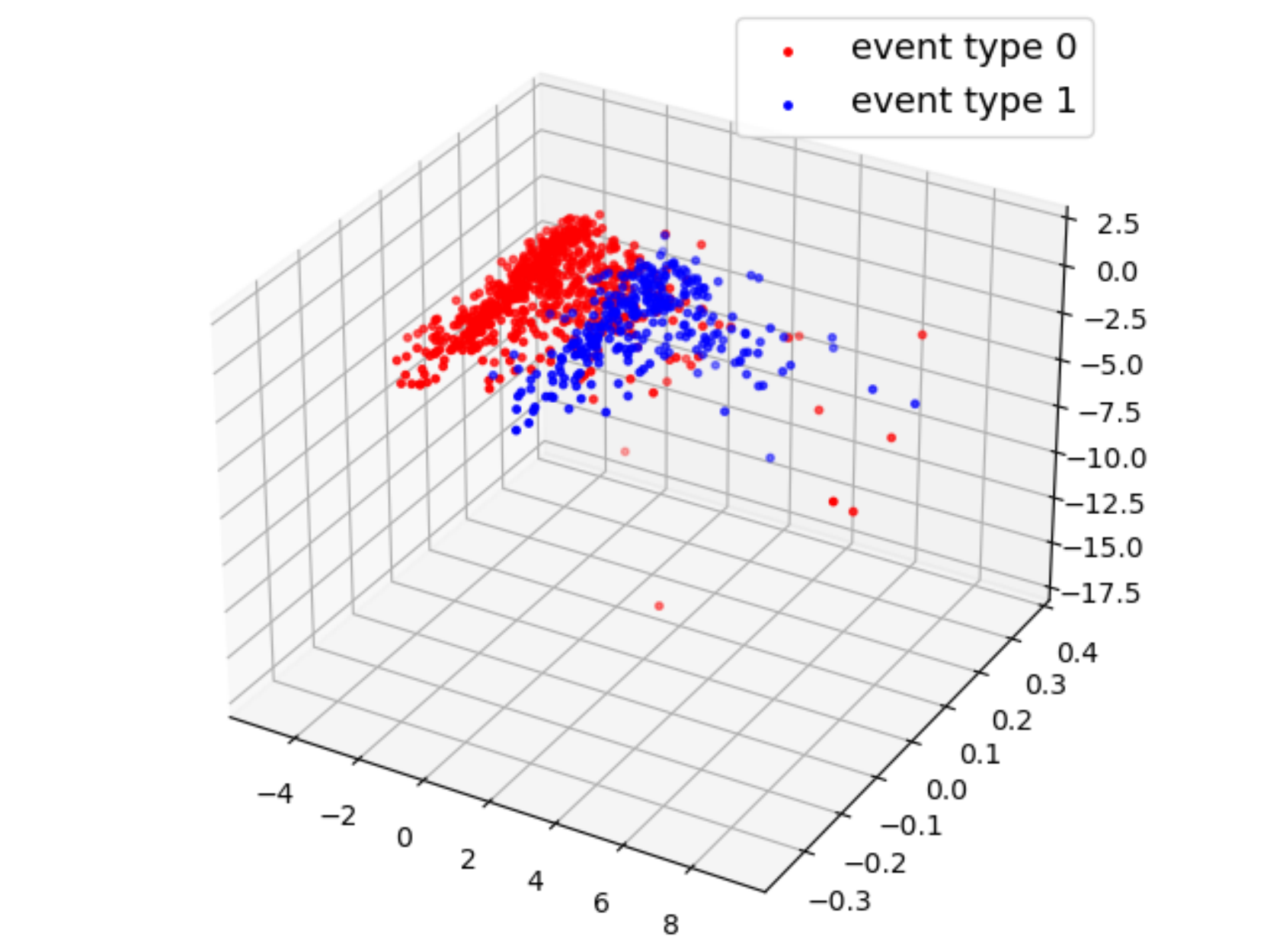} }}%
    \:
    \subfloat[\centering ]{{\includegraphics[width=0.22\textwidth]{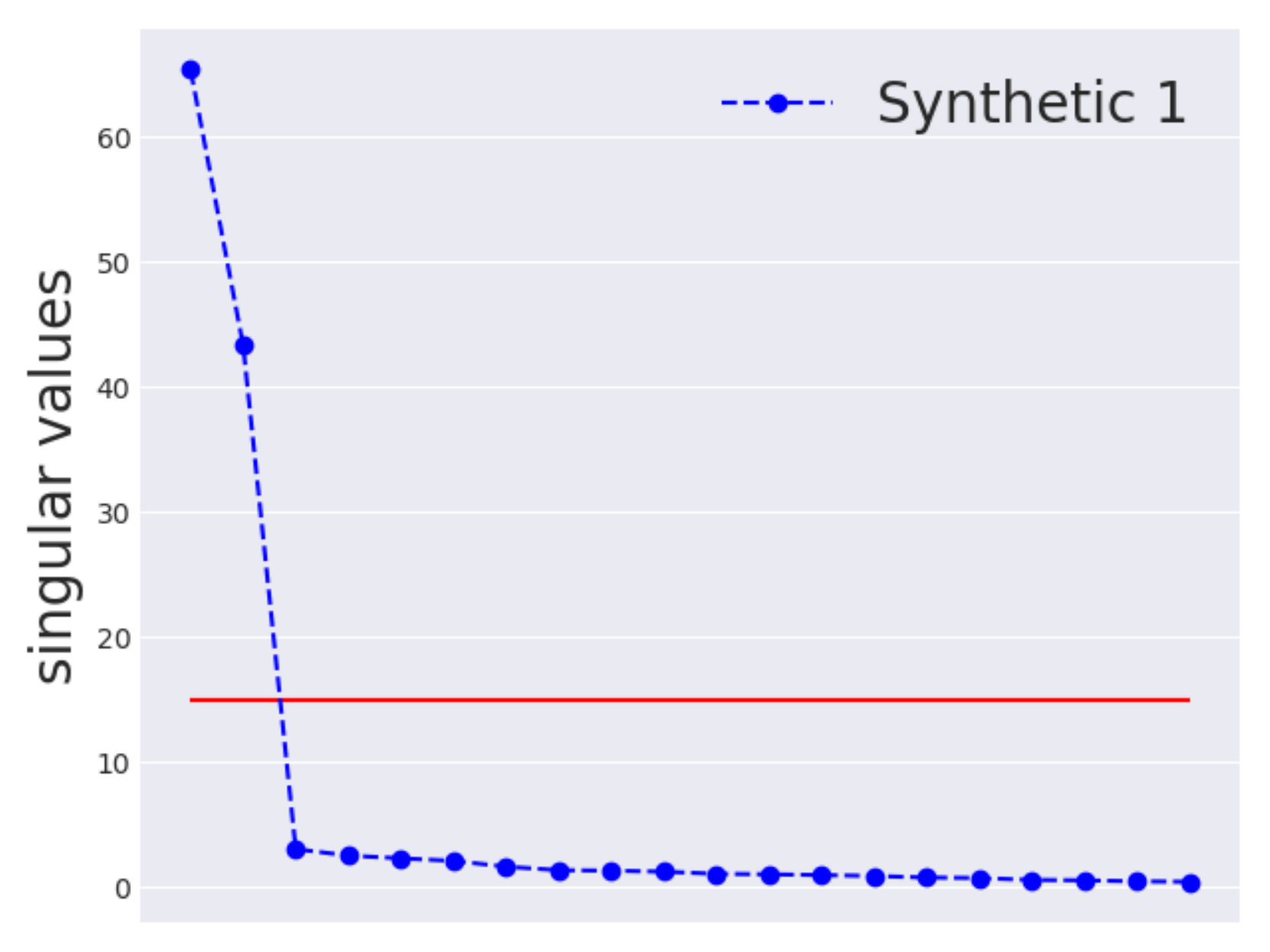} }}%
    \:
    \subfloat[\centering ]{{\includegraphics[width=0.22\textwidth]{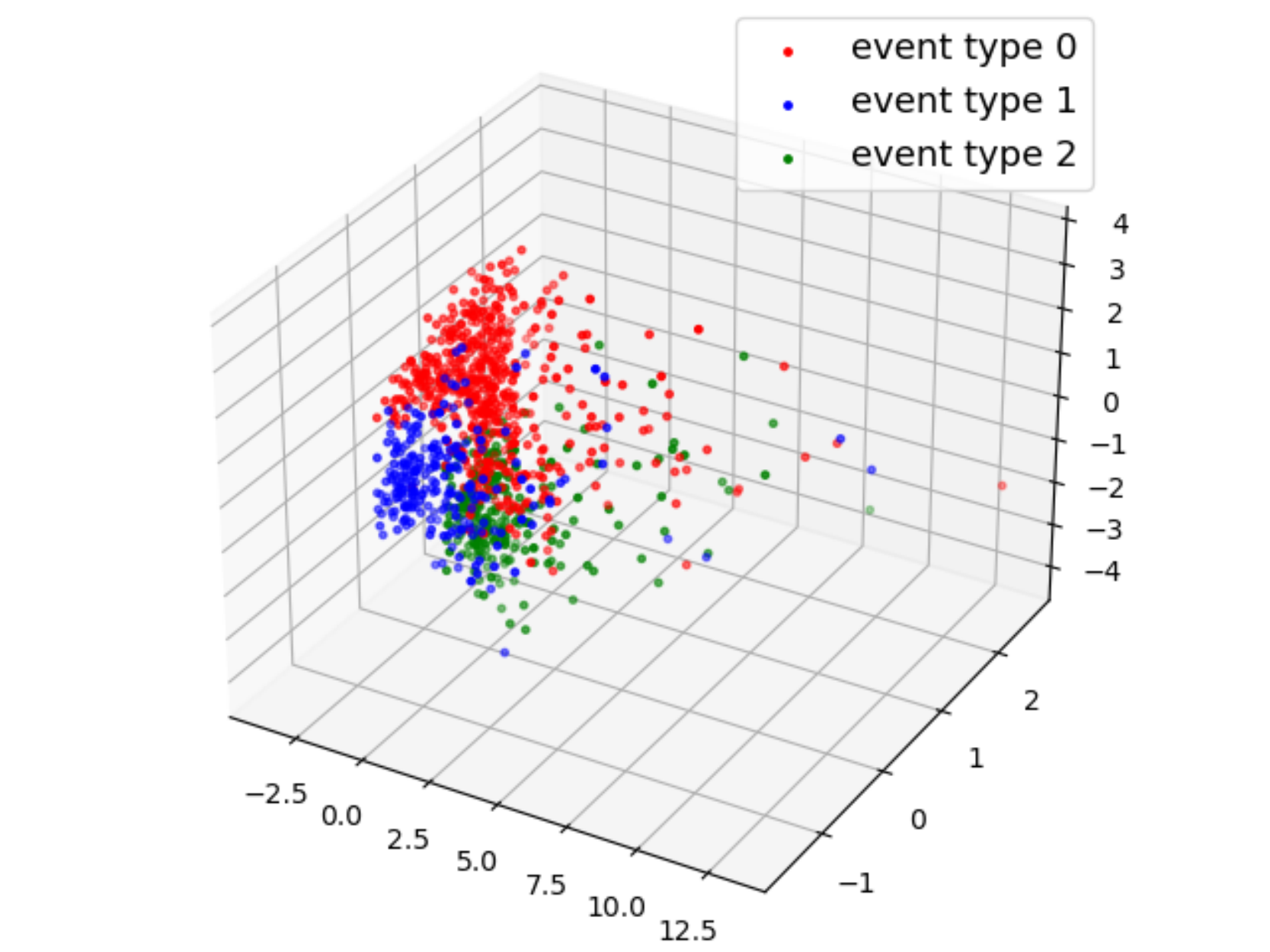} }}%
    \:
    \subfloat[\centering ]{{\includegraphics[width=0.22\textwidth]{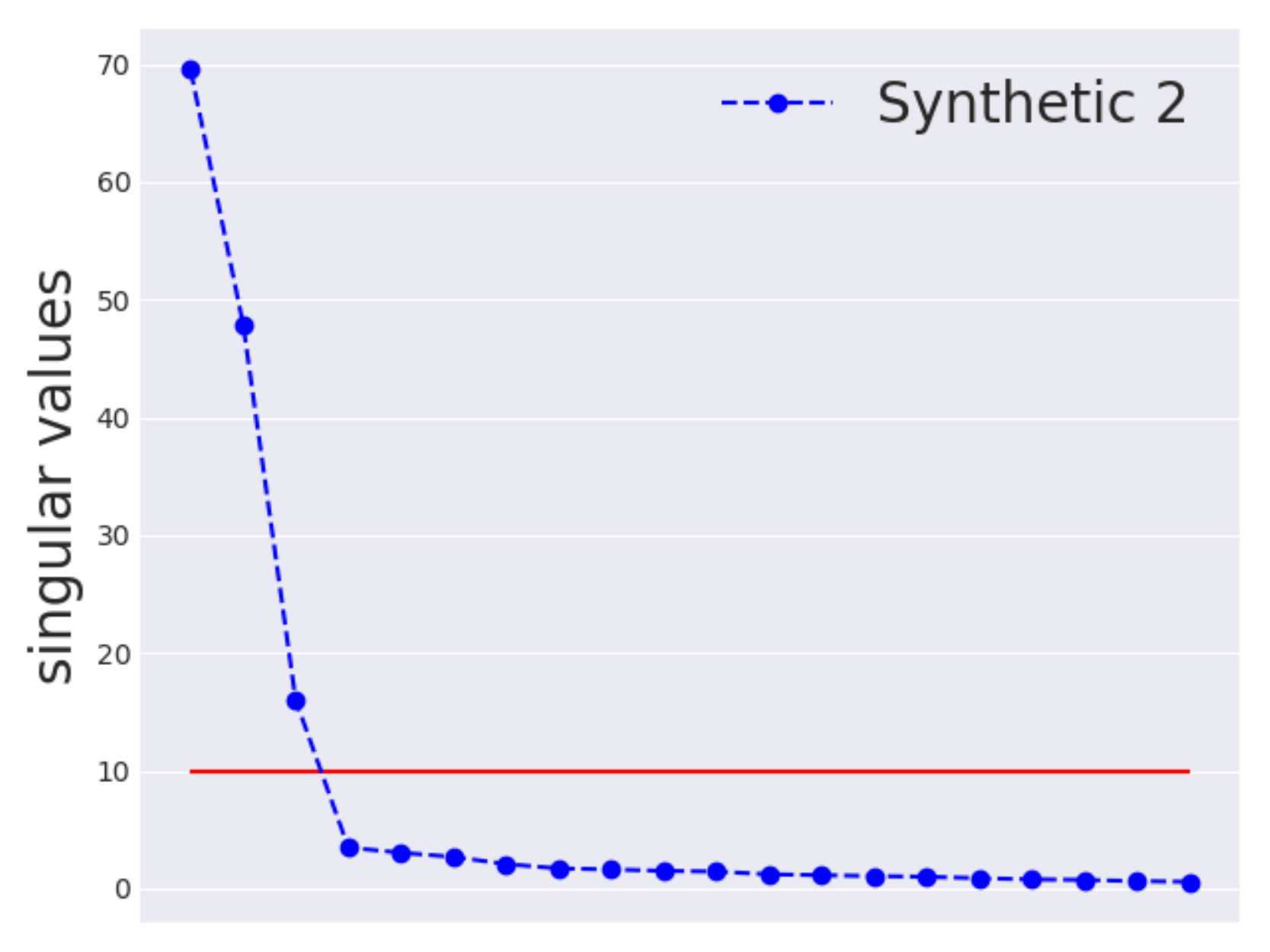} }}%
    \caption[Latent variable SVD]{The singular value decomposition results on latent variable $z$ from VNTPP-L. We use three axes, each with top three singular values, and project the latent variables. The top left and bottom left present the projection results for the Synthetic 1 and Synthetic 2 datasets, respectively. The top right and bottom right show the singular values of latent variables for the Synthetic 1 and Synthetic 2 datasets, respectively.}%
    \label{fig:svd}%
\end{figure}

\begin{figure}[t!]%
    \centering
    \subfloat[\centering ]{{\includegraphics[width=0.22\textwidth]{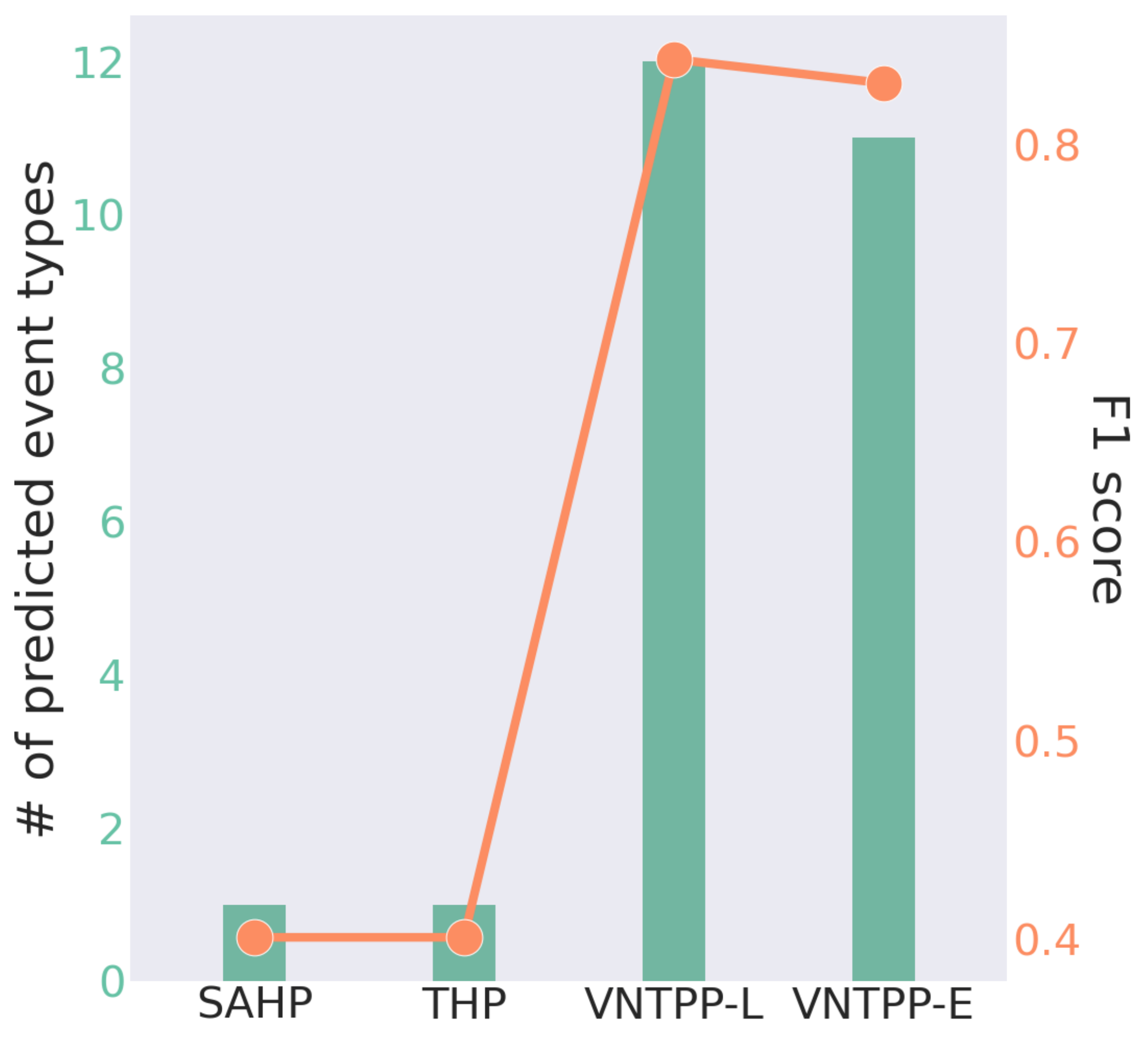} }}%
    \:
    \subfloat[\centering ]{{\includegraphics[width=0.22\textwidth]{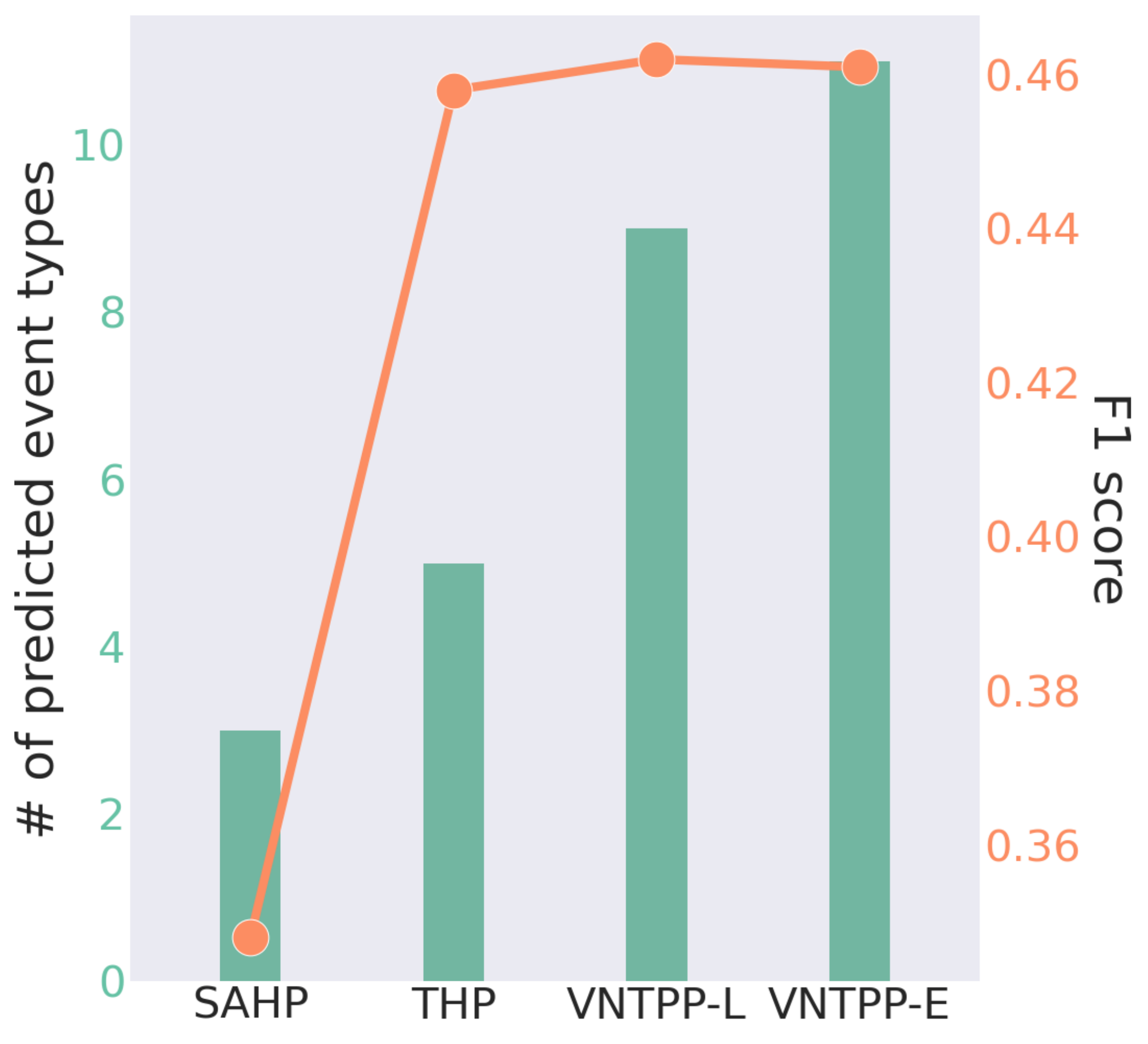} }}%
    \caption[The number of correctly predicted event types on MIMIC-II and StackOverflow datasets]{The number of correctly predicted event types on MIMIC-II (left) and StackOverflow (right) datasets. MIMIC-II has 75 types of events, and StackOverflow has 22 types. As shown in this figure, VNTPP is able to predict diverse event types and achieves better performances in event predictions simultaneously.}%
    \label{fig:diverity}%
\end{figure}

\begin{figure*}[t!]%
    \centering
    \subfloat[\centering ]{{\includegraphics[width=0.59\textwidth]{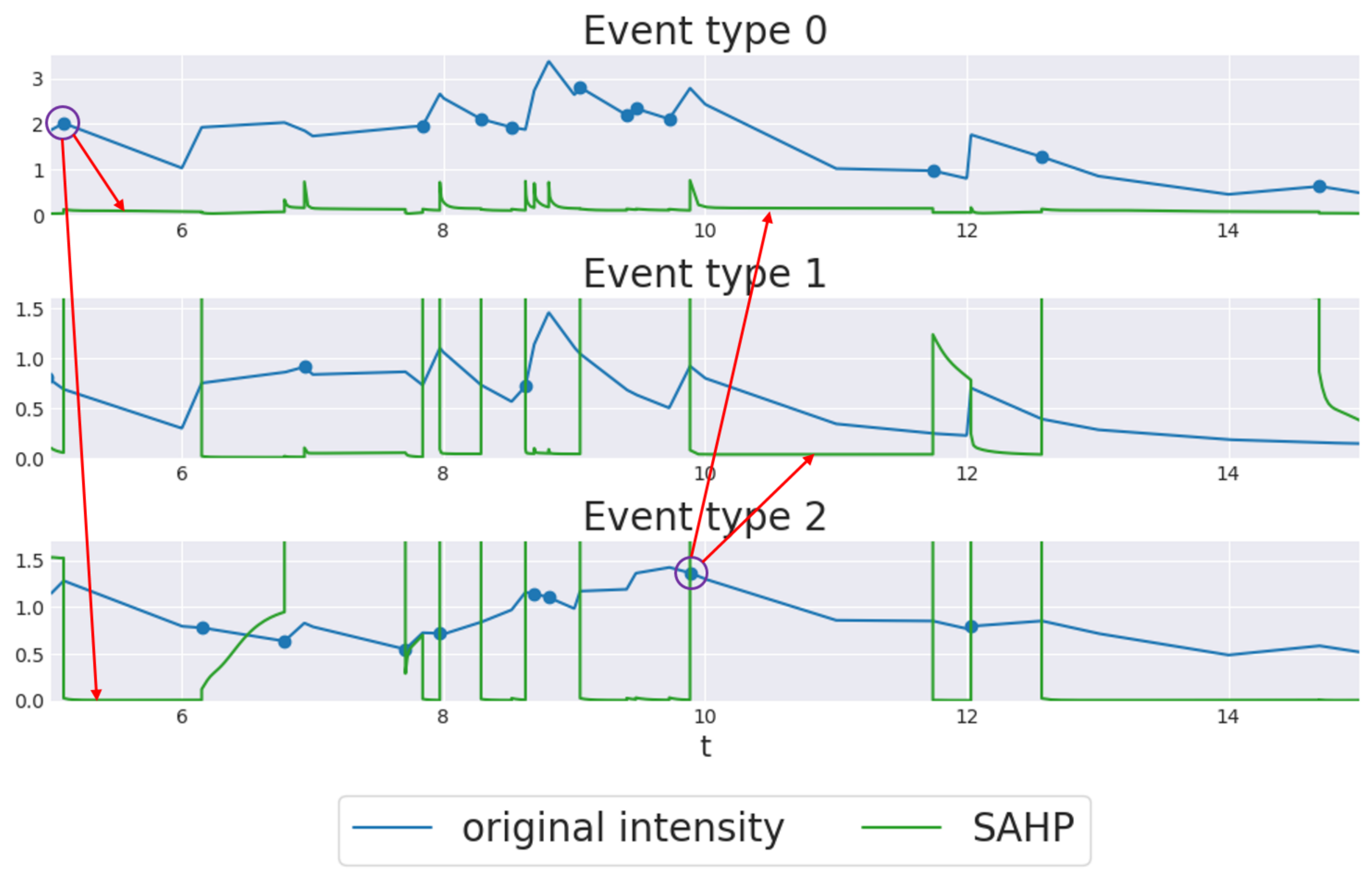} }}%
    \:
    \subfloat[\centering ]{{\includegraphics[width=0.59\textwidth]{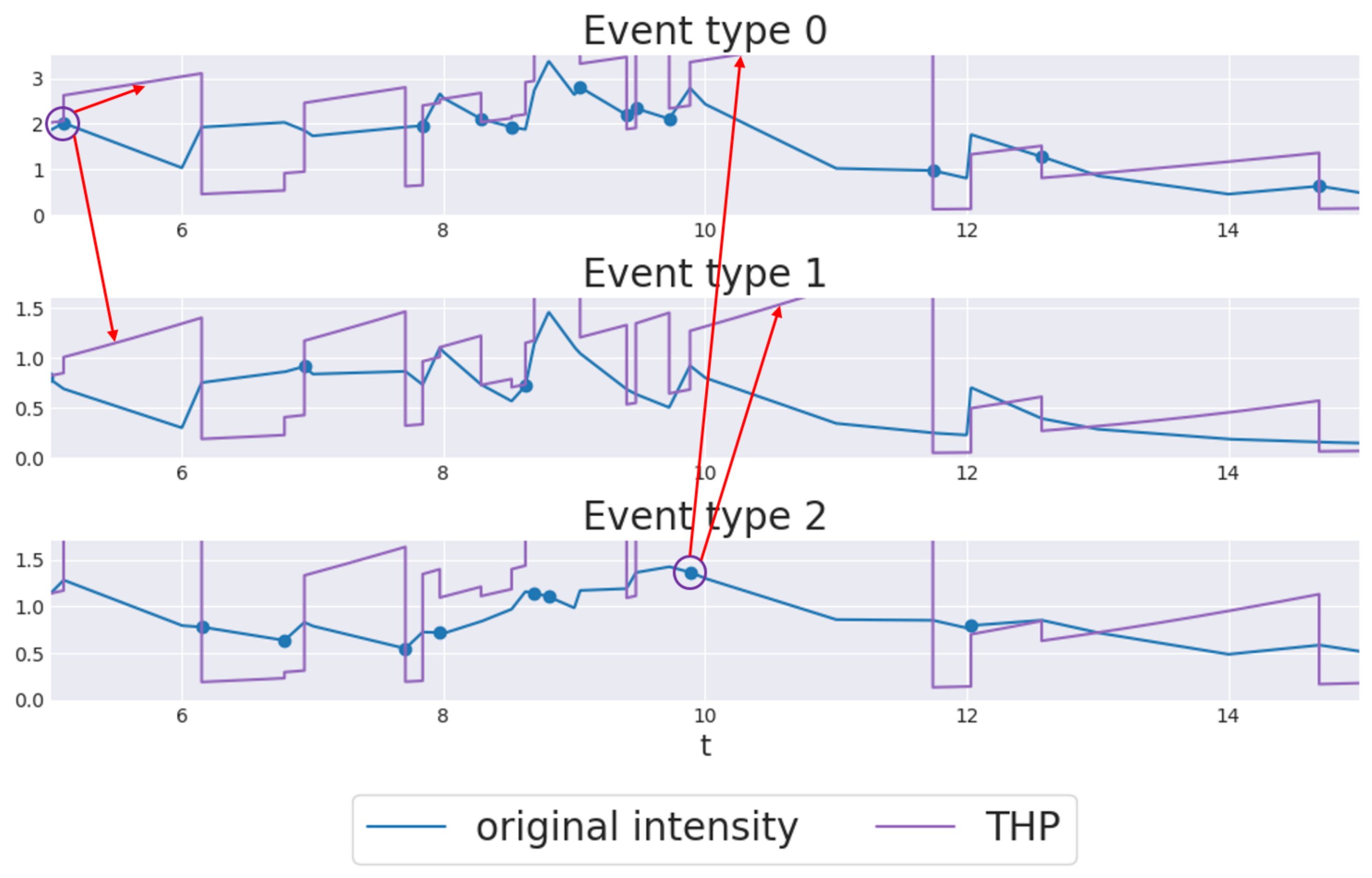} }}%
    \:
    \subfloat[\centering ]{{\includegraphics[width=0.59\textwidth]{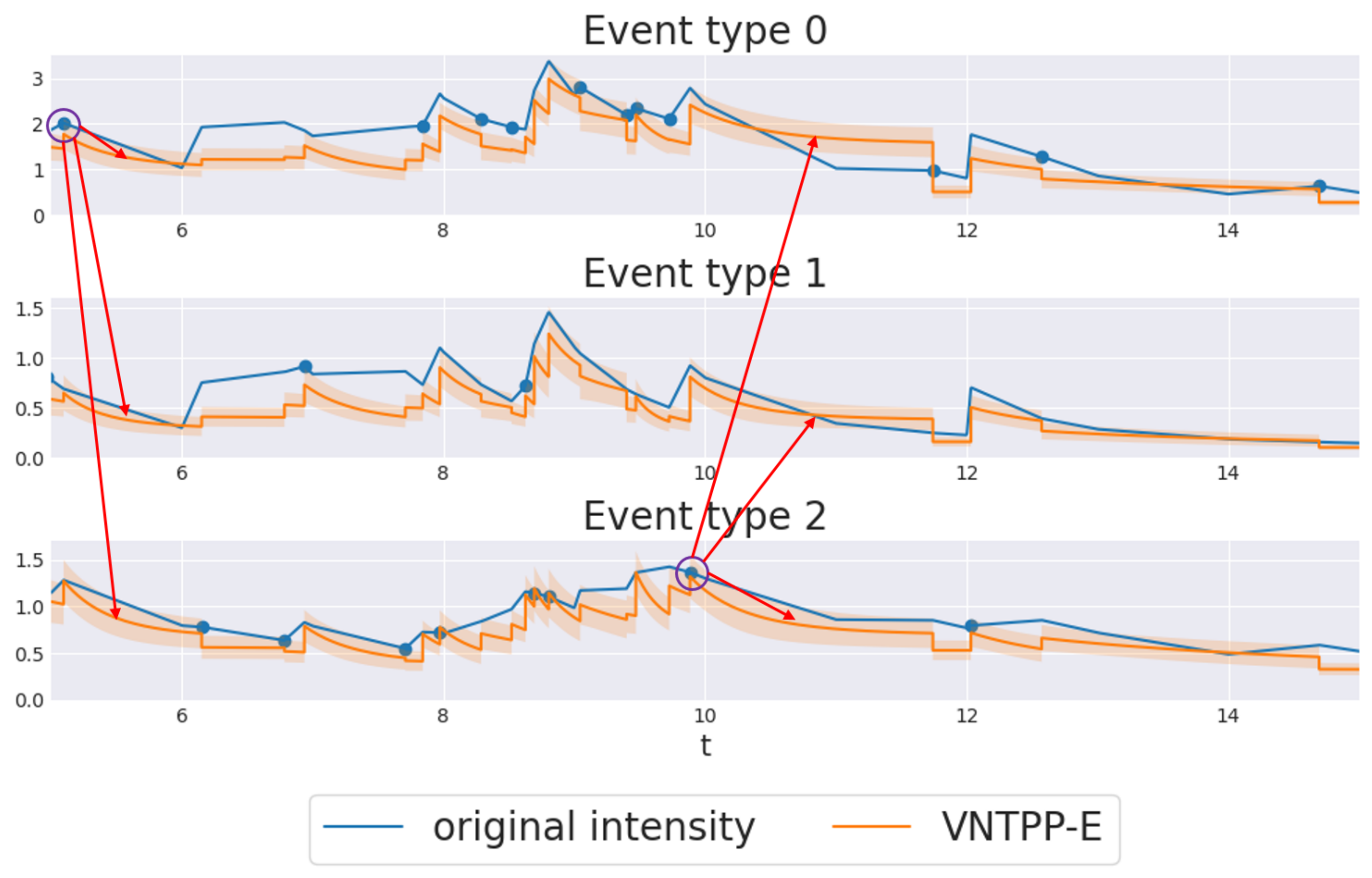} }}%
    \caption[Detailed comparison for intensity estimation]{Detailed comparison for intensity estimation. The purple circles are the event occurrences and the red arrows are the reactions for the event occurrences in the purple circles}%
    \label{fig:event_interaction}%
\end{figure*}

\subsection{Training Details}
In the inference and generative networks, the hidden dimension of the self-attention layer is 64, $d_k=d_v=16$, and the number of heads is 4. The number of layers is 2, the dropout rate is $0.1$. In addition, the prior distribution of $z$ is the unit Gaussian distribution, and the number of epoch is 300. We choose the number of dimensions of the latent variable $z$ from $[10, 20, 40]$ and the learning rate from $[5\times 10^{-4}, 1\times 10^{-4}, 1\times 10^{-3}]$. We use Adam \cite{DBLP:journals/corr/KingmaB14} optimizer with the PyTorch framework. VNTPP-L and VNTPP-E use linearly and exponentially decaying intensity functions with respect to time respectively.

\subsection{Event Type and Time Prediction}
Table \ref{tab:f1_result} and \ref{tab:rmse_result} show the event type and time prediction results. VNTPP achieves the highest performances on type F1 scores and time RMSEs. This verifies the effectiveness of our proposed variational approach compared to SAHP and THP, which are based on the self-attention mechanism. Even when additional linear layers are not used to predict event type and time, and VNTPP only utilizes the computed intensity functions with numerical approximations, VNTPP achieves better performances than THP. On the synthetic datasets, although HP-EK and HP-GK achieve good performances for event type predictions, the models are restricted with fixed functional kernel forms. Therefore, the Hawkes processes are unstable on long and complicate sequences. On the other hand, VNTPP shows competitive performances on both the real-word and synthetic datasets.

\subsection{Conditional Intensity Function Evaluation}
We are able to compute conditional intensity functions when generating the synthetic datasets. Therefore, we compare the intensity functions with the estimated intensity functions from SAHP, THP and VNTPP. Table \ref{tab:intensity_result} provides the quantitative evaluations of the estimated intensity functions. VNTPP predicts the intensity functions more accurately than SAHP and THP. In addition, Fig \ref{fig:intensity_function} illustrates the comparison of intensity functions from the Synthetic 2 dataset. The intensity functions are computed using latent variable $z$ and $z$ is sampled under Gaussian distribution with $z_\mu$ and $z_\sigma$. Accordingly, we draw 100 latent variables and obtain 100 intensity functions for each time. The solid line for VNTPP is the mean of the intensity function, and the shaded area represents the $\pm 3\sigma$ areas of the generated intensity functions. The original intensity functions of the two figures in Fig \ref{fig:intensity_function} are the same. We split the figures to compare VNTPP with SAHP and THP separately since the scales of estimated intensity functions are different. Although we use the sampled latent variables to compute intensity functions, the variations in the estimated intensity functions are not large. In addition, the estimated intensity functions for VNTPP are closer to the original intensity functions than SAHP and THP on Synthetic 1 and 2. The intensity functions from SAHP and THP have large values on several timestamps. On the other hand, VNTPP follows the shape of the original intensity functions better. In addition, VNTPP can generate different intensity functions for each event. This means that VNTPP recognizes different event types from the input data. This experiment indicates VNTPP can predict intensity functions more stably and accurately than baselines.

\subsection{Latent Variable}
To show that latent variables well represent the history and next event type in VNTPP, we conduct a singular value decomposition (SVD) and choose the three axes that have the top three largest singular values. The result is described in Fig \ref{fig:svd}. The top left and top right are the projection results on the three axes and the singular values on the Synthetic 1 dataset. Synthetic 1 has two types of event, so there are two colors on the projection plot. The bottom left and right are results for the Synthetic 2 dataset, which has three types of events. In the two projection plots, the latent variables are automatically separated according to the event types. No other loss or regularizers are used to separate the latent variables. The model automatically learns the different representations for each event. In addition, in two singular value plots, the number of large singular values are the same as the number of event types in the two datasets. The synthetic datasets are generated with the fixed kernel forms of the Hawkes processes, so this is a natural result. However, if we use real-world datasets, the number of large singular values is more than the number of event types since real-world datasets are not generated with specific kernel forms or fixed point processes. Thus, we need more axes to draw the projection plots of latent variables. In these results, we show that the latent variables well describes the attributes of input history and next event type.

Since we train the distribution of the latent variable, VNTPP is able to estimate the distributions of intensity functions. Fig \ref{fig:density_contour} shows the difference between deterministic model and stochastic model. The baselines based on deep neural network are deterministic so cannot estimate the distributions of the intensity functions. On the other hand, our model can estimate the mean and variance of intensity functions by sampling the latent variables repeatedly. In this way, our model can handle the data uncertainty, i.e., out-of-distribution data and noisy data. In addition, the latent variables from VNTPP are well clustered automatically and centered on a specific point for each event type. On the other hand, the latent variables from THP are disorderly distributed and it is difficult to interpret the latent variables.

\subsection{Event Diversity}
Fig \ref{fig:diverity} shows the correctly predicted event diversity for each model on MIMIC-II and StackOverflow. Since the other datasets contain only three types of events, to compare the capabilities of the models, we only use two datasets which have many rare event types. VNTPP is able to generate the most diverse event types in neural network based models. The neural network can easily be overfitted to a few dominant event types on the training dataset. Therefore, some models converge just to the one most dominant event. However, VNTPP can predict diverse event types and is not easily overfitted to a specific event type. In the training step, we sample latent variables under the approximate posterior distribution, and in the inference step, we use the mean of the latent variables to predict the events in deterministic way. In this way, sampling latent variables prevents the model from overfitting on the training step, and VNTPP can predict the events in a stable way on the inference step.

\subsection{Detailed Comparison for Intensity Estimation}
Fig \ref{fig:event_interaction} describes the detailed comparison for the intensity function estimation. We set the same ranges for x-axis and y-axis in three figures. The shape of intensity functions are changed as soon as one of the three events arrives. The purple circles indicate the event occurrences and the red arrows are the reactions for the event arrivals in the purple circles. VNTPP properly predicts the shape changes of the intensity functions, but other models are unstable and some predicted intensities are too large. As for SAHP, most of the shapes of the intensity functions between two events are straight lines parellel to the x-axis. In addition, some of the intensities have very large values or small values. As for THP, the model learns the increasing functions as time goes on, instead of decaying property. In addition, the shapes of the intensity function between two events are almost same. In other words, the model overfits to one specific straight line and different histories are not discriminated on this model. Like SAHP, some of the intensity values are extremely large so some part of the graphs are outside of the fixed range. On the other hand, VNTPP-E learns different decaying shapes for each different history and the estimation accuracy are highest among three neural network based models. The model reacts properly as soon as one of the events arrives. In addition, there is no extremely high value, so the model is able to predict the intensity functions stably.

\section{Conclusion}
We suggest a novel temporal point process based on a deep neural network. We introduce the latent variable $z$ to predict conditional intensity functions and the intensity functions are used to estimate next event type and time. VNTPP is able to better learn the hidden representations of input histories. Training the distributions of latent variables instead of point estimation has positive effects on learning neural temporal point processes on complex and noisy time series datasets that contain randomness and variability due to rare event types. In addition, we can measure the uncertainty of intensity functions, and the latent variables of VNTPP are interpretable. As a result, VNTPP outperforms other baselines on event time, event type, and intensity estimation.

\bibliography{aaai22}

\end{document}